\documentclass[a4paper,10pt]{article}
\usepackage[margin = 1.0in]{geometry}

\usepackage{latexsym}
\usepackage{amsmath}
\usepackage{amssymb}
\usepackage{amsthm}
\usepackage{thmtools}
\usepackage{epsfig}
\usepackage{ulem} 
\usepackage{hyperref}
\usepackage{enumerate}
\usepackage{cleveref}
\usepackage{listings}
\usepackage{xfrac}

\usepackage[color, leftbars]{changebar}

\usepackage[ruled,vlined]{algorithm2e}
\usepackage[linewidth=1.2pt,linecolor=black,nobreak=true]{mdframed}
\mdfsetup{frametitlealignment=\center}
\usepackage{changepage}

\newcommand{\remove}[1]{}

\def\eps{\ensuremath{\epsilon}}

\def\P{\ensuremath{\mathcal{P}}}

\newcommand{\PP}{\mathbb{P}}

\newcommand{\F}{\ensuremath{\mathcal{F}}}
\newcommand{\G}{\ensuremath{\mathcal{G}}}

\renewcommand{\P}{\ensuremath{\mathcal{P}}}
\newcommand{\A}{\ensuremath{\mathcal{A}}}
\newcommand{\B}{\ensuremath{\mathcal{B}}}

\newcommand{\myproof}[1][\textit{Proof}]{\textit{#1. }}

\newcommand{\abs}[1]{ \ensuremath{\left| #1 \right|}  }

\newcommand{\round}[1]{ \ensuremath{\left( #1 \right)}  }

\newtheorem{theorem}{Theorem}

\newtheorem{lemma}[theorem]{Lemma}

\newtheorem{claim}[theorem]{Claim}

\newcommand{\etal}{{et al}.\@ }
\DeclareMathOperator{\vcdim}{\mathrm VC-dim}

\newtheorem{thmx}{Theorem}

\newcommand{\mystatef}[1]{
	\begin{mdframed}[hidealllines=true, innermargin = 0.2in, outermargin = 0.2in]
		\vspace{0.1in}
		#1
	\end{mdframed}
}

\title{Optimal Approximations Made Easy}
\date{}

\author{M\'onika Csik\'os  \\
	{\small Universit\'e Gustave Eiffel, LIGM,  Equipe A3SI, ESIEE Paris,} \\ {\small Cit\'e Descartes, 2 boulevard Blaise Pascal, 93162 Noisy-le-Grand Cedex, France.}\\{\small
	Email: monika.csikos@esiee.fr}\\~
\and 
Nabil H. Mustafa \\
	{\small Université Sorbonne Paris Nord,
Laboratoire LIPN CNRS 7030,
Institut Galilée,
} \\ {\small 99 avenue Jean-Baptiste Clément
93430 Villetaneuse, France.}\\{\small
	Email: nabil.mustafa@univ-paris13.fr }}

\begin{document}

	\maketitle

	\begin{abstract}
\noindent		The fundamental result of 	Li, Long, and Srinivasan~\cite{LiLS-sample-complexity-learning-S01} on approximations of  set
		systems has become a key tool across several communities such as learning theory, algorithms, computational geometry, combinatorics, and data analysis.\\ \\
			The goal of this paper is to give a modular, self-contained, intuitive proof of this result for finite set systems.
		The only ingredient we  assume is the standard Chernoff's concentration bound. 	This makes the proof accessible to a wider audience, readers not familiar with techniques from statistical learning theory, and makes it possible to be covered in a single self-contained lecture in a geometry, algorithms
		or combinatorics course.
	\end{abstract}


\noindent
Funding: The work of the authors has been supported by the grant ANR ADDS (ANR-19-CE48-0005).

	\section{Introduction}

	Given a finite set system $(X, \F)$, our goal is to construct a small set $A \subseteq X$ such
	that each set of  $\F$ is `well-approximated' by $A$.
	Research on such approximations started in the 1950s, with
	random sampling being the key   tool for showing their existence.
	A breakthrough in the study of approximations dates back to 1971 when Vapnik and Chervonenkis  studied set systems with finite VC-dimension \cite{VC71}.
	The \textit{VC-dimension} of $(X,\F)$, denoted by $\vcdim(X,\F)$,
	is the size of the largest $Y \subseteq X$ for which $\F|_Y = 2^Y$,
	where $\F|_{Y} = \left\{ Y \cap S \colon S \in \F \right\}$.
	Since then, the notion of approximations has become a fundamental structure across
	several communities---learning theory, statistics, combinatorics and algorithms (see \cite{nabilbook}).

	\paragraph{Relative $\left(\eps, \delta\right)$-approximations.}
		Given a set system $(X, \F)$ with $n = |X|$
		and parameters $0<\eps,\delta <1$, a set $A$ of size $t$ is a \textit{relative $(\eps,\delta)$-approximation} for $(X,\F)$ if for all $S \in \F$,
		\begin{align*}
		\abs{ \frac{|S| }{n} - \frac{|A \cap S|}{t}} \leq \delta\cdot \max \left\{ \frac{|S|}{n}, \eps  \right\}, \quad \text{or equivalently,} \quad
		|A \cap S| =  \frac{|S| \, t}{n} \pm \delta t \max \left\{ \frac{|S|}{n}, \eps \right\}.
		\end{align*}

	In this paper, we study guarantees for relative $\left(\eps, \delta\right)$-approximations obtained by random sampling. In particular, given a set $X$ we say that $A$ is a uniform random sample of $X$ of size $t$ if $A$ is selected uniformly at random from the ${n \choose t}$ subsets of $X$ of size $t$.


	A basic guarantee follows immediately from Chernoff's bound (for completeness, we give the standard proof in the Appendix).
		\begin{thmx} \label{thm:chernoffboundapprox}
			Let $X$ be a set of $n$ elements and $A$ be a uniform random sample of $X$ of size $t$.
			Then for any $S \subseteq X$ and $\eta > 0$,
			\begin{equation*} 
			\PP \left[ |A \cap S| \notin\round{ \frac{|S| t}{n} - \eta,~ \frac{|S| t}{n} + \eta} \right]
			\leq
			2 \, \exp \left( - \frac{\eta^2 n}{2|S|t + \eta n} \right).
			\end{equation*}
		In particular, setting $\eta = \delta t \max \left\{ \frac{|S|}{n}, \eps \right\}$,
		a uniform random sample $A$ of size $t$ fails to be a relative $(\eps, \delta)$-approximation for  a fixed
		$S \in \F$ with probability at most $2 \exp \left( -\frac{\eps \delta^2 \, t}{3} \right)$.
		\end{thmx}

		\Cref{thm:chernoffboundapprox} in conjunction with the union
		bound gives the following upper-bound on relative $(\eps, \delta)$-approximation sizes for \emph{any} finite set system (the detailed proof is presented in the Appendix).
		\begin{theorem} \label{sec:generalprobabilisticconstructions:thm:basicrelativeapproximation}
			Let $\left(X, \F\right)$ be a finite set system and
			$0 < \eps, \delta, \gamma < 1$ be given parameters. Then for any integer $t \geq \frac{3}{\eps \delta^2} \ln \frac{2 |\F|}{\gamma}$, a uniform random
			sample $A \subseteq X$ of size $t$
			is a relative $(\eps, \delta)$-approximation for $\F$ with probability at least $1 - \gamma$.
		\end{theorem}

	This paper addresses the following influential result
	of Li, Long, and Srinivasan \cite{LiLS-sample-complexity-learning-S01}, described as `the pinnacle of a long sequence of papers'
	in~\cite[Section 7.4]{Har-peled:2011:GAA:2031416}.\footnote{The original result was stated using the notion of  $(\eps, \delta)$-samples, but they are asymptotically equivalent:  an $(\eps,\delta)$-sample is a relative $(\eps,4\delta)$-approximation and a relative $(\eps, \delta)$-approximation is an $(\eps, \delta)$-sample; see~\cite{Har-PeledS11-relative-approximation-geometry}.}
		\begin{theorem}[\cite{LiLS-sample-complexity-learning-S01}]\label{maintheorem}
			There exists an absolute constant $c$ such that the following holds.
			Let $(X, \F)$ be a set system such that $|\F|_Y| \leq \left( e |Y|/d \right)^d$ for all $Y \subseteq X$ with $|Y| \geq d$, and  let $0 < \delta, \eps, \gamma < 1/2$ be given parameters. Then for any integer 
			$$
			t \geq \frac{c}{\eps \delta^2} \cdot \left( d \ln \frac{1}{\eps} + \ln \frac{1}{\gamma} \right),
			$$
			 a uniform random sample $A \subseteq X$ of size $t$ is a relative $(\eps,\delta)$-approximation for $(X,\F)$ with probability at least $1 - \gamma$.
		\end{theorem}
		\textbf{Remarks.}
		\begin{enumerate}
			\item Note that by the Sauer-Shelah lemma,  $\vcdim \left( X, \F \right) \leq d$ implies
		that $|\F|_Y| \leq \round{ e|Y|/d}^d$ for any $Y \subseteq X$ (see e.g., \cite[Lemma 10.2.5]{M02}). Thus, Theorem
		\ref{maintheorem} also applies to set systems with $\vcdim \left( X, \F \right) \leq d$.
        \item This bound is asymptotically tight~\cite{LiLS-sample-complexity-learning-S01} and immediately implies
        many other approximation bounds such as $\eps$-approximations (Vapnik and Chervonenkis~\cite{VC71}, Talagrand \cite{T94}),
        sensitive $\eps$-approximations (Br{\"o}nnimann \etal\cite{BCM93}), and $\eps$-nets (Haussler and Welzl~\cite{HW87}, Koml\'os \etal\cite{KPW92}).
    	\end{enumerate}

\medskip

	The original proof of Theorem \ref{maintheorem} uses two techniques:

	\begin{description}
    \item[Symmetrization.] To prove
	that a   random sample $A$ satisfies the required properties,
	one takes another random sample $G$, sometimes called a `ghost sample'. Properties of $A$ are then proven by comparing it with $G$. Note that $G$ is not used
	in the algorithm or its construction---it is solely a method of analysis, a `thought experiment' of sorts.


	\item[Chaining.] The idea is to analyze the interaction of the sets in $\F$ with a random sample
	by partitioning each $S \in \F$ into a logarithmic number of smaller sets, each   belonging to a distinct `level'. The number of sets increase with increasing level while the size of each set decreases. The overall sum turns out to be a geometric series, which then  gives the optimal bounds~\cite{KT59, talagrand2016upper}.

\end{description}

What makes  the proof of \Cref{maintheorem} in~\cite{LiLS-sample-complexity-learning-S01} difficult
is that it  combines chaining and symmetrization intricately. All the tail bounds are stated
in their `symmetrized' forms and symmetrization is carried through the entire proof. It is not an easy proof to
explain to undergraduate or even graduate students in computer science, as it is difficult to see what is really going on in terms of the significance
and intuition of these two ideas. In fact, even the proofs of simpler statements involving
just symmetrization,  as given in textbooks\footnote{Also used in teaching; to pick two arbitrary examples, see {\color{blue} \href{https://web.stanford.edu/class/cs229t/2015}{here} }for an example from the perspective of statistics/learning and {\color{blue} \href{https://www.ti.inf.ethz.ch/ew/courses/CG12/index.html}{here} }from the algorithmic side.}---e.g., see \cite{KV94, DGL96, M99, Chazelle:2000:DMR:507108, M02, AB09, Har-peled:2011:GAA:2031416, DBLP:books/wi/AlonS92,nabilbook}---often come with the caveat that the idea is  ingenious  but difficult to understand intuitively (e.g., ``one might
	be tempted to believe that it works by some magic''~\cite[Section 10.2]{M02}).

%

\subsection*{Our Results.}

This work is an attempt to improve this state of affairs.  
We show that in fact one can separate the roles of chaining and symmetrization,
giving two  separate statements which together immediately imply \Cref{maintheorem}.
The role of symmetrization is to get a bound on relative $(\eps, \delta)$-approximations that is independent of $|\F|$ (but contains an additional factor of $\log\frac 1 \delta$):
\mystatef{
\begin{restatable}{theorem}{relativeapproxwithlog} \label{thm:relativeapproxwithlog}
				There exists an absolute constant $c_1$ such that the following holds.
			Let $(X, \F)$ be a set system such that $|\F|_Y| \leq \left( e |Y|/d \right)^d$ for all $Y \subseteq X$, $|Y| \geq d$, and  let $0 < \delta, \eps, \gamma <1/2$ be given parameters. Then for any integer 
			$$
			t \geq \frac{c_1}{\eps \delta^2} \cdot \left( d \ln \frac{1}{\eps \delta} + \ln \frac{1}{\gamma} \right),
			$$
			a uniform random sample $A \subseteq X$ of size $t$ is a relative $(\eps,\delta)$-approximation for $(X,\F)$ with probability at least $1 - \gamma$.
	\end{restatable}
}
\textbf{Remark.}  The proof of \Cref{thm:relativeapproxwithlog} is standard using symmetrization. For completeness, we  present a different, folklore proof at the end of the paper (\Cref{sec:symmetrization}), which in fact shows that symmetrization is not really necessary for finite set systems\footnote{This is typically the case in its use in algorithms, computational geometry, combinatorics. The infinite case can usually be reduced	 to the finite case by a sufficiently fine grid, see~\cite{MatousekWW93-discrepancy-approx-VC}.}  and can be replaced by a   more intuitive argument
that makes it obvious, pedagogically, why the bound is independent of $|\F|$.

On the other hand, the role of chaining is to get rid of logarithmic factors that arise when applying union bound, by more carefully analyzing the failure probability for a collection of events.
The key observation is that  \Cref{thm:chernoffboundapprox} provides a bound on the probability of failure  for a set $S \in \F$ which  \emph{decreases} as the size of $S$ decreases. One can take advantage of this by partitioning each $S \in \F$ into a logarithmic number of smaller sets, each belonging to a distinct level, such that the levels strike a proper balance---the number of sets (arising from partitioning every $S \in \F$) increase each level, but their size  across levels decreases geometrically. This way one gets an improved bound by applying the union bound separately to sets of different levels.

The resulting bound is captured in the next statement (it removes the factor of $\log \frac 1 \delta$, but depends on $|\F|$):
\mystatef{
 \begin{restatable}{theorem}{chaining} \label{thm:chaining}
				There exists an absolute constant $c_2$ such that the following holds.
			Let $(X, \F)$ be a set system such that $|\F|_Y| \leq \left( e |Y|/d \right)^d$ for all $Y \subseteq X$, $|Y| \geq d$, and  let $0 < \delta, \eps, \gamma < 1/2$ be given parameters. Then for any integer 
			$$
			t \geq c_2 \ \max \left\{ \frac{1}{\eps \delta} \ln \frac{ |\F|}{\gamma},~~  \frac{1}{\eps \delta^2} \ln   \left( \frac{1}{\eps^d\gamma} \right) \right\},
			$$
			a uniform random sample $A \subseteq X$ of size $t$ is a relative $(\eps,\delta)$-approximation for $(X,\F)$ with probability at least $1 - \gamma$.
\end{restatable}
}
The proof of \Cref{thm:chaining} is given in \Cref{sec:chaining}.

The above two statements immediately imply a proof of \Cref{maintheorem}: given $(X, \F)$, apply \Cref{thm:relativeapproxwithlog}
to get a set $A_1 \subseteq X$ of size
$O \left( \frac{1}{\eps \delta^2} \ln \frac{1}{\eps^d \delta^d \gamma} \right)$, which is a relative $(\eps, \frac{\delta}{3})$-approximation of $\F$ with probability at least $1-\frac \gamma 2$.
Now apply  \Cref{thm:chaining} to $\F|_{A_1}$ to get $A_2 \subseteq A_1$ of size
$$ O \left( \max \left\{ \frac{1}{\eps \delta} \ln \frac{  \left( \frac{e}{d \eps \delta^2} \ln \frac{1}{\eps^d \delta^d \gamma} \right)^d }{\gamma},~~  \frac{1}{\eps \delta^2} \ln   \left( \frac{1}{\eps^d\gamma} \right) \right\} \right)
= 	O \left(		\frac{1}{\eps \delta^2} \cdot \left( d \ln \frac{1}{\eps} + \ln \frac{1}{\gamma} \right) \right),$$
which is a relative $(\eps, \frac{\delta}{3})$-approximation of $\F|_{A_1}$
with probability at least $1-\frac \gamma 2$.
Thus $A_2$ is a relative $(\eps, \delta)$-approximation of $\F$ of the required size with probability at least $1- \gamma $.


	\section{Proof of Theorem \ref{thm:chaining}}
		\label{sec:chaining}


	Let  $n = |X|$ and $t = |A|$.
	We use the following  consequence of  \Cref{thm:relativeapproxwithlog} (better bounds
exist~\cite{Haussler92spherepacking, M16}; however the one derived below suffices for our needs):

	\mystatef{
	\begin{lemma} \label{lemma:packing-lemma}
		There is an absolute constant $c_3$ such that the following holds.
		Let  $\alpha \geq 2$ and let $\P \subseteq \F$ be an $\alpha$-packing of $\F$; that is, for any   $S,S' \in \P$, the symmetric difference of $S$ and $S'$, denoted by $\Delta(S,S')$, has size at least $\alpha$.
		Then
		$
		|\P|  \leq   \round{ \frac{c_3n}{ \alpha} }^{2d}.
		$
	\end{lemma}
	\myproof
				Let $\G = \left\{ \Delta \left(S, S'\right) \colon S, S' \in \P\right\}$.
 By \Cref{thm:relativeapproxwithlog}
			there exists a relative $(\frac{\alpha}{n}, \frac{1}{2})$-approximation $A'$ for $\G$ of size
			$$
			|A'| = \frac{  c_1}{\frac{\alpha}{n} \cdot \frac 1 4} \round{ d \ln\frac{2n}{\alpha} + \ln\frac{2n}{\alpha}}
			\leq
			\frac{8c_1dn}{\alpha} \cdot \ln \frac{2n}{\alpha}
			\leq
			\frac{8c_1 d n^2}{\alpha^2},
			$$
			where we set $\gamma = \frac{\alpha}{2n}$ (note that we could set any positive value for $\gamma$ as we only use the \emph{existence} of such approximations).
			Then for any $S, S' \in \P$, we get
			$$
			\left| \Delta \left(S, S'\right) \cap A' \right|
			\geq
			\frac{\left|\Delta \left(S, S'\right)\right| \, |A'|}{n} -\frac{|A'|}{2} \cdot  \max\left\{\frac{\left|\Delta \left(S, S'\right)\right| }{n}, \frac{\alpha}{n}\right\}
			=
			\frac{1}{2} \cdot  \frac{\left|\Delta \left(S, S'\right)\right| \, |A'|}{n}
			> 0.$$
			This implies that $A' \cap S \neq A' \cap S'$ for any $S, S' \in \P$, and so we have that $|\P| = |\P|_{A'}|$. Finally, we use that $\P \subset \F$ and thus $|\P| = |\P|_{A'}| \leq |\F|_{A'}| \leq \left( \frac{8ec_1 n^2}{\alpha^2} \right)^d = \left( \frac{\sqrt{8ec_1}n}{\alpha} \right)^{2d}$. Setting $c_3 = \sqrt{8ec_1}$ concludes the proof.
		\qed
 }


	Set 	$k = \left\lceil\log \frac{1}{\delta} \right\rceil$ and for   $i \in [0, k]$, let
	$\P_i$ be a \emph{maximal} $\frac{\eps n}{2^i}$-packing of $\F$ and set $\P_{k+1} = \F$.
	For any $S \in \P_{i+1} \setminus \P_i$ there exists a set  $ F_S \in \P_i $ such that $ |\Delta(S,F_S)| < \frac {\eps n} {2^i}$.
	Define
	\begin{align*}
	\A_i = \left\{ S \setminus F_S \colon S \in \P_{i+1} \setminus \P_i \right\} \qquad \text{and} \qquad
	\B_i = \left\{ F_S \setminus S \colon S \in \P_{i+1} \setminus \P_i \right\}.
	\end{align*}
	Lemma \ref{lemma:packing-lemma} implies that
	\begin{align*}
	|\A_i|, |\B_i| \leq |\P_{i+1}| \leq \round{\frac{c_3 \cdot 2^{i}}{\eps}}^{2d}.
	\end{align*}
	\begin{claim}  \label{claimepsiapproxforaibi}
		Let $\eps_i =    \sqrt{ (i+1)/2^{i} } \,  \eps $. With probability  $1-\gamma$,   $A$ is simultaneously
		\vspace{-1em}
		\begin{enumerate}[(i)]
			\item a relative $(\eps, \delta)$-approximation for $\A_k \cup \B_k$, and
			\item a relative $(\eps_i, \delta)$-approximation for $\A_i \cup \B_i$ for all $i \in [0, k-1]$, and
			\item a relative $(\eps, \delta)$-approximation for $\P_0$.
		\end{enumerate}
	\end{claim}
\myproof
		$(i)$ Each set in $\A_k \cup \B_k$ has size less than  $\frac{\eps n}{2^k} \leq \eps n \delta \leq \eps n$. Therefore, we apply \Cref{thm:chernoffboundapprox} with $\eta = \delta t \eps$  and take the union bound over $|\A_k \cup \B_k| \leq 2|\F|$ sets which gives that for a large-enough value of $c_2$, $A$ fails to be an $(\eps, \delta)$-approximation for $\A_k \cup \B_k$ with probability at most
		\begin{align*}
		2|\F| \cdot 2 \exp \left( - \frac{\delta^2 t^2 \eps^2 \cdot n}{2 \eps n \delta \cdot t +\delta   t \eps \cdot n } \right)
		= 2|\F| \cdot 2 \exp \left( - \frac{\delta \eps t}{3} \right) \leq \frac{\gamma}{3}.
		\end{align*}

		$(ii)$ For a fixed $S \in \A_i \cup \B_i$, we have $|S| \leq \frac{\eps n }{2^i} \leq \eps_i n$. Thus, applying \Cref{thm:chernoffboundapprox} with $\eta = \delta  t \eps_i$ implies that the probability of failure for a fixed set $S \in \A_i \cup \B_i$ is at most
		\begin{align*}
		2 \exp \round{
			{-} \frac{\delta^2 t^2 \eps_i^2   n }{2|S|t {+} \delta \eps_i t n }
		}
		&\leq
		2 \exp \round{
			- \frac{\delta^2  t \eps^2 {(i{+}1)}/{2^{i}}      }{{2\eps}/{2^i} {+} \delta \eps \sqrt{{(i {+}1)}/{2^{i}}}   }
		}
		\leq
		2 \exp \round{
			- \frac{\eps \delta^2 t (i{+}1)}{4}
		}.
		\end{align*}
		Hence, by the union bound, the overall probability of failure is at most
		\begin{align*}
		\sum\limits_{i=0}^{k-1}
		\left|\A_i \cup \B_i\right|
		\cdot 2 \exp \round{
			- \frac{\eps \delta^2 t (i{+}1)}{4}
		}
		&\leq
		\sum\limits_{i=0}^{k-1}
		2 \round{
			\frac{c_3 \cdot 2^{i}}{\eps}}^{2d}
		2 \left( \eps^d \gamma \right)^{c_2 (i+1)/4}
		\leq
		\gamma \sum\limits_{i=0}^{k-1}
		\frac{4 \round{
			c_3 \cdot 2^{i-1}}^{2d}}{
		 {2^{(d+1)c_2 (i+1)/4}}}\\
		&
		\leq \gamma \, \sum\limits_{i=1}^{\infty}
		\frac{1}{5^i}
		\leq \frac{\gamma}{3},
		\end{align*}
		for  $c_2 = 8\log_2 c_3 + 18 \geq 8\round{\log_2 c_3 + \frac{\log_2(5)}{2d+2}+1}$.\\
		$(iii)$ Since $|\P_0| \leq \round{\frac{c_3}{\eps}}^{2d}$,  \Cref{sec:generalprobabilisticconstructions:thm:basicrelativeapproximation} implies that this failure probability is at most $\frac{\gamma}{3}$
		if $t \geq \frac{3}{\eps \delta^2} \ln \frac{2 \left({c_3}/{\eps}\right)^{2d}}{\gamma/3}$.
\qed

	Observe that for any set  $S \in  \F$, there exists a set $S_{k} \in \P_{k}$, with $A_{k} = S \setminus S_{k} \in \A_{k}$ and $B_{k} = S_{k} \setminus S \in \B_{k}$, such that $S = \left( S_{k} \setminus B_{k} \right) \cup A_{k}$. Similarly, one can express $S_k$ in terms of $S_{k-1} \in \P_{k-1}$, $A_{k-1} \in \A_{k-1}$, $B_{k-1} \in \B_{k-1}$ and so on until we reach $S_0 \in \P_0$.  Thus
	using Claim \ref{claimepsiapproxforaibi},   with probability at least $1- \gamma $,
	\begin{align*}
	&\bigg| \frac{|S|}{n} -  \frac{|A \cap S|}{t}  \bigg|
	=
	 \abs{ \frac{|S_{k} |}{n}- \frac{| B_{k} |}{n}+ \frac{|A_{k}|}{n} - \round{\frac{|A \cap S_{k}|}{t}- \frac{|A \cap  B_{k} |}{t}+ \frac{|A \cap A_{k}|}{t}}} \\
	&~~ \stackrel{(i)}{\leq}
	\abs{ \frac{|S_{k} |}{n}-\frac{|A \cap S_{k}|}{t}}+ \delta \max\left\{\eps, \frac{|A_{k}|}{n}\right\}+ \delta \max\left\{\eps, \frac{|B_{k}|}{n}\right\}
	=
	\abs{ \frac{|S_{k} |}{n}-\frac{|A \cap S_{k}|}{t}}+ 2 \delta \eps  \  \leq \
	\cdots  \\
	&~~ {\stackrel{(ii)}{\leq} }
	\abs{ \frac{|S_{0} |}{n}-\frac{|A \cap S_{0}|}{t}}+ 2\delta\sum_{j=0}^{k-1} \eps_j + 2 \delta \eps \\
	&~~{\stackrel{(iii)}{\leq}}
	\delta \max \left\{ \eps, \frac{|S_0|}{n}  \right\}+ 14 \delta \eps
	\leq
	\delta \frac{|S|}{n}+ 16 \delta \eps
	\leq
	2 \delta\max \left\{ \frac{|S|}{n}, 16 \eps  \right\},
	\end{align*}
	where the second-last step uses the fact that $|S_0| \leq |S| +  \sum\limits_{j=0}^{k} |B_i|
	\leq     |S| + \sum\limits_{j=0}^{\infty} \frac{\eps n}{2^j}
	\leq  |S| +   2\eps n$.

	Therefore, $A$ is a relative $(16\eps, 2\delta)$-approximation of $\F$
	with probability at least $1-\gamma$. Repeating the same arguments with $\delta' = \delta/2$ and $\eps' = \eps/16$, we get  a relative $\left(\eps,  \delta \right)$-approximation of $\F$, as required.
	\qed

\section{Proof of \Cref{thm:relativeapproxwithlog}} \label{sec:symmetrization}

 The proof uses an argument similar to the discrepancy-based argument
 used for $\eps$-approximations~\cite{MatousekWW93-discrepancy-approx-VC}, though it is somewhat simpler as it does not need discrepancy, and it applies to the more general notion of a relative $(\eps, \delta)$-approximation.

To see the intuition, observe
		that since $|\F| \leq (e |X|/d)^d$, the bound of \Cref{sec:generalprobabilisticconstructions:thm:basicrelativeapproximation} depends
		only on  $|X|$---in particular that a random sample $A_1 \subseteq X$ of size  $O \left( \frac{1}{\eps \delta^2} \ln |X|^d \right)  = O \left( \frac{d}{\eps \delta^2} \ln |X| \right)$ is a relative $(\eps, \delta)$-approximation.
		The size of $A_1$ is much smaller than that of $X$
		and so applying \Cref{sec:generalprobabilisticconstructions:thm:basicrelativeapproximation} again to $\F|_{A_1}$ gives a relative $(\eps, \delta)$-approximation $A_2 \subseteq A_1$
		for $\F|_{A_1}$, with
		$$ |A_2| = O \left( \frac{1}{\eps \delta^2} \ln |A_1|^d \right) = O \left( \frac{d}{\eps \delta^2} \ln \left( \frac{d}{\eps \delta^2} \ln |X| \right) \right)
		= O \left( \frac{d}{\eps \delta^2} \ln \frac{d}{\eps \delta} + \frac{d}{\eps \delta^2} \ln \ln |X| \right).$$
		The  size of $A_2$ is again much smaller than that of $A_1$. Furthermore, it follows immediately from the definition
		of relative $(\eps, \delta)$-approximations that
		$A_2$ is a relative $\left(\eps, 3\delta \right)$-approximation for $\F$.
		With each successive application of \Cref{sec:generalprobabilisticconstructions:thm:basicrelativeapproximation}, the size of
		the set decreases rapidly, while the error of approximation increases only linearly, giving the required bound that is independent of $|\F|$.

	Now we turn to the formal proof of \Cref{thm:relativeapproxwithlog}. Let $T \left( \eps, \delta, \gamma \right)$ be the smallest integer
	such that a uniform random sample of size at least $T \left( \eps, \delta, \gamma \right)$ from $X$ is a relative $(\eps, \delta)$-approximation for $\F$
	with probability at least $1-\gamma$. Further define $\delta_{0} = 0$ and  $\delta_i = \frac{3^{i-1}}{\sqrt{|X|}}$ for $i = 1, \dots, \left\lceil \frac 1 2 \log_3(\sqrt{|X|})\right\rceil + 1$.
	We prove that for all $i$, for all $ \epsilon, \gamma \in (0,1/2)$ and for all $\delta \in (\delta_{i-1}, \delta_i]$, it holds that
	$
	T \left(\eps, \delta, \gamma\right) \leq \frac{c_1}{\eps \delta^2} \cdot \left( d \ln \frac{1}{\eps \delta} + \ln \frac{1}{\gamma} \right),
	$ 
	which is equivalent to the desired statement. The proof is by induction on $i$.\\
	\textbf{Base case ($i=1$):} When $\delta \in (0, \delta_0]$, we have   $|X| \leq \frac{1}{\delta^2} $ and thus $T\left(\eps, \delta, \gamma\right)$
	is upper-bounded as required for any $\epsilon, \gamma \in (0,1/2)$.\\
	\textbf{Inductive hypothesis ($j \leq i$):} Assume that the statement holds for all $j \leq i$, that is, for any $\delta \in (0, \delta_i]$ and $\eps, \gamma \in (0,1/2)$, we have 
	$
	T \left(\eps, \delta, \gamma\right) \leq \frac{c_1}{\eps \delta^2} \cdot \left( d \ln \frac{1}{\eps \delta} + \ln \frac{1}{\gamma} \right).
	$ \\
	\textbf{Inductive step ($i \to i+1$):} Let $\delta \in \left(\delta_i, \delta_{i+1}\right]$. Since $\frac \delta 3 \in
	(0, \delta_i]$, the inductive hypothesis gives that a  random sample $A' \subseteq X$ of size $T \left( \eps, \frac{\delta}{3}, \frac{\gamma}{2} \right) \leq \frac{9c_1}{\eps \delta^2} \cdot \left( d \ln \frac{3}{\eps \delta} + \ln \frac{2}{\gamma} \right)$,
	is a relative $\left(\eps, \frac{\delta}{3}\right)$-approximation for $\F$ with probability at least $1  - \frac{\gamma}{2}$.
	By Theorem~\ref{sec:generalprobabilisticconstructions:thm:basicrelativeapproximation}, a uniform random sample $A$ of $A'$ of size 
	$$
		\frac{3}{\eps (\delta/3)^2} \ln \frac{2 \, |\F|_{A'}|}{(\gamma/2)}
	$$ is a relative $\left(\eps, \frac{\delta}{3}\right)$-approximation for $\F|_{A'}$ with probability $1-\frac{\gamma}{2}$.
	Thus $A$ is a uniform random sample of $X$ that is a relative $(\eps, \delta)$-approximation
	for $\F$ with probability at least  $ 1- \gamma$, implying the recurrence
	$$ T \left( \eps, \delta, \gamma \right) \leq |A|
	= \frac{3}{\eps (\delta/3)^2} \ln \frac{2 \, |\F|_{A'}|}{(\gamma/2)}
	\leq \frac{27}{\eps \delta^2} \ln \left( \frac{4}{\gamma} \, \left( \frac{e \, T \left( \eps, \frac{\delta}{3}, \frac{\gamma}{2} \right)}{d} \right)^d \right).$$
	The required bound on $T\left( \eps, \delta, \gamma\right)$ now follows by the inductive hypothesis.
	As $\left(1+\frac{1}{d} \ln \frac{2}{\gamma}\right)^d \leq \frac{2}{\gamma}$,
	\begin{align*}
		\frac{27}{\eps \delta^2} \ln \left( \frac{4}{\gamma} \left( \frac{e \,  \frac{9 \, c_1}{\eps \delta^2} \left( d \ln \frac{3}{\eps \delta} + \ln \frac{2}{\gamma} \right) }{d} \right)^d \right)
		\leq \frac{27}{\eps \delta^2} \ln \left(  \frac{4}{\gamma}   \left( \frac{e \, 27\, c_1}{\eps^2 \delta^3} \right)^d \left( 1 + \frac{1}{d} \ln \frac{2}{\gamma} \right)^d  \right)
		\leq \frac{c_1}{\eps \delta^2}  \ln \frac{1}{ \left(\eps \delta\right)^d \gamma},
	\end{align*}
	for any constant $c_1\geq 318$, which concludes the proof of \Cref{thm:relativeapproxwithlog}.
	\qed

\paragraph{Acknowledgement.} We would like to thank the anonymous reviewers for their valuable comments which improved this paper.

	\bibliographystyle{alpha}
	\bibliography{simpleproofLLS_bib}

 \section*{Appendix}

\paragraph{Proof of \Cref{thm:chernoffboundapprox}.}

		  $|A\cap S|$ follows hypergeometric distribution with expectation $\frac{|S|t}{n}$.
Thus, we can apply
the standard Chernoff's tail estimate \cite[Theorem 21.6 and Section 21.5]{frieze2016introduction} to get
		\begin{align*}
			\PP \left[ |A \cap S| \notin\round{ \frac{|S| t}{n} - \eta,~ \frac{|S| t}{n} + \eta} \right]
			&=
			\PP \left[ |A \cap S| \leq \frac{|S| t}{n} - \eta \right]
			+
			\PP \left[ |A \cap S| \geq  \frac{|S| t}{n} + \eta \right]\\
			&\leq
			\exp\round{-\frac{\eta^2}{2\round{|S|t/n - \eta/3}}}
			+
			\exp\round{-\frac{\eta^2}{2\round{|S|t/n + \eta/3}}}\\
			&\leq
			2\exp\round{-\frac{n\eta^2}{2|S|t + n\eta}}.
		\end{align*}

	\qed

\paragraph{Proof of \Cref{sec:generalprobabilisticconstructions:thm:basicrelativeapproximation}.}
			By \Cref{thm:chernoffboundapprox}, a uniform random sample $A$ of size $t$ fails to be a relative $(\eps, \delta)$-approximation for  a fixed
		$S \in \F$ with probability at most $2 \exp \left( -\frac{\eps \delta^2 \, t}{3} \right)$. By the union bound,
		\begin{align*}
			\PP \left[\exists S \in \F \text{ s.t. } |A \cap S| \notin\round{ \frac{|S| t}{n} - \delta t \max \left\{ \frac{|S|}{n}, \eps \right\},~ \frac{|S| t}{n} + \delta t \max \left\{ \frac{|S|}{n}, \eps \right\}} \right]
			\leq
			|\F| \cdot 2 \exp \left( -\frac{\eps \delta^2 \, t}{3} \right)
			\leq
			\gamma.
		\end{align*}
		Therefore, with probability at least $1-\gamma$, $A$ is a relative $(\eps, \delta)$-approximation for any set $S \in \F$.
\qed

\end{document}